\documentclass[iicol,pdflatex,sn-mathphys-num]{sn-jnl}

\usepackage{graphicx}%
\usepackage{float}%
\usepackage{multirow}%
\usepackage{amsmath,amssymb,amsfonts}%
\usepackage{amsthm}%
\usepackage{mathrsfs}%
\usepackage[title]{appendix}%
\usepackage{xcolor}%
\usepackage{textcomp}%
\usepackage{manyfoot}%
\usepackage{booktabs}%
\usepackage{algorithm}%
\usepackage{algorithmicx}%
\usepackage{algpseudocode}%
\usepackage{listings}%

\theoremstyle{thmstyleone}%
\theoremstyle{thmstyletwo}%

\theoremstyle{thmstylethree}%

\begin{document}

\title[Article Title]{Optimized Detection and Classification on GTRSB: Advancing Traffic Sign Recognition with Convolutional Neural Networks}

\author*[1]{\fnm{Dhruv} \sur{Toshniwal}}\email{dhruvt@bu.edu}
\author*[2]{\fnm{Saurabh} \sur{Loya}}\email{saurabh.loya@utah.edu}
\author[3]{\fnm{Anuj} \sur{Khot}}\email{anujkhot41@gmail.com}
\author[3]{\fnm{Yash} \sur{Marda}}\email{mardayash001@gmail.com}

\affil[1]{\orgdiv{Department of Computer Science}, \orgname{Boston University}, \orgaddress{\street{111 Cummington Mall}, \city{Boston}, \postcode{02215}, \state{MA}, \country{USA}}}

\affil[2]{\orgdiv{Kahlert School of Computing}, \orgname{The University of Utah}, \orgaddress{\street{50 Central Campus Dr., Room 3190}, \city{Salt Lake City}, \postcode{84112}, \state{UT}, \country{USA}}}

\affil[3]{\orgdiv{School of Computer Engineering and Technology}, \orgname{Dr. Vishwanath Karad MIT World Peace University}, \orgaddress{\street{Survey No, 124, Paud Rd}, \city{Pune}, \postcode{411038}, \state{Maharashtra}, \country{India}}}

\abstract{
In the rapidly evolving landscape of transportation, the proliferation of automobiles has made road traffic more complex, necessitating advanced vision-assisted technologies for enhanced safety and navigation. These technologies are imperative for providing critical traffic sign information, influencing driver behavior, and supporting vehicle control, especially for drivers with disabilities and in the burgeoning field of autonomous vehicles. Traffic sign detection and recognition have emerged as key areas of research due to their essential roles in ensuring road safety and compliance with traffic regulations. Traditional computer vision methods have faced challenges in achieving optimal accuracy and speed due to real-world variabilities. However, the advent of deep learning and Convolutional Neural Networks (CNNs) has revolutionized this domain, offering solutions that significantly surpass previous capabilities in terms of speed and reliability. This paper presents an innovative approach leveraging CNNs that achieves an accuracy of nearly 96\%, highlighting the potential for even greater precision through advanced localization techniques. Our findings not only contribute to the ongoing advancement of traffic sign recognition technology but also underscore the critical impact of these developments on road safety and the future of autonomous driving.}

\keywords{CNN, Image Classification and Recognition, Traffic Sign Analysis, Machine Learning.}

\maketitle
\section{Introduction}\label{sec1}

In the contemporary landscape, marked by the ascendancy of electric vehicles and automation, it becomes important to carefully address human errors. Artificial Intelligence (AI) is particularly refined to circumvent the replication of such errors. Within the realm of automobile automation, the automation of traffic sign detection stands out as a pivotal mechanism for preempting rule infractions, thereby respecting law. Such a system not only mitigates the likelihood of traffic violations but also paves the way for streamlined enforcement and maintenance processes, encouraging adherence to traffic regulations among the general populace.

To pioneer in this domain, we propose the study of a neural network-based Traffic Sign Recognition System (TSRS) designed for rigorous training and empirical validation in real-world settings. TSRSs are crucial for a plethora of applications, including autonomous driving, traffic monitoring, enhancement of driver safety, maintenance of road networks, and traffic scene analysis. At its core, a TSRS addresses two intertwined aspects: Traffic Sign Detection (TSD) and Traffic Sign Recognition (TSR)\cite{8}. Traffic signs, integral to road infrastructure, are crafted to capture the attention of both pedestrians and drivers promptly, providing guidance and warnings irrespective of the time of day. The distinct design of traffic signs, characterized by simplistic geometries and standardized color schemes, ostensibly simplifies the task of detection and recognition. However, the quest for a resilient, real-time TSRS is fraught with challenges stemming from environmental variables such as scale disparities, adverse viewing angles, motion blur, color degradation, obstructions, and varying lighting conditions.

Compounding these technical challenges is the existence of over 300 traffic sign categories, as delineated by the Vienna Convention on Road Traffic, a treaty endorsed by 63 countries. Despite the treaty’s widespread adoption, minor discrepancies in traffic sign designs across different nations introduce additional complexities to the recognition process. A proficient TSRS must, therefore, demonstrate robustness against such diversities. The quintessential function of traffic signs—to conspicuously convey crucial information pertinent to the road segment in question—aims to preemptively alert drivers to potential hazards and encourage adherence to designated speed limits, thereby fostering a safer driving environment.

\section{Literature Review}\label{sec2}

Numerous studies have advanced the field of real-time Traffic Sign Recognition (TSR) and classification, employing various methodologies to enhance accuracy and reliability. This section reviews significant contributions to the TSR domain, highlighting the methodologies and findings of key research efforts.

In the work of Akshata V.S and Subarna Panda \cite{1}, the TSR system is divided into three main stages: image segmentation, traffic sign detection, and classification based on the input image. Their methodology leverages a color enhancement technique to isolate red regions within the image, facilitating the subsequent detection, classification, and recognition processes using Convolutional Neural Networks (CNN). The primary aim of their research is to accurately classify, recognize, and identify German traffic signs under various conditions. This is achieved through a CNN architecture composed of neurons with learnable weights and biases. The process begins with loading data from the German Traffic Sign Benchmark, followed by data exploration and visualization. The final steps involve designing the model architecture for training and testing, applying the model to new images, and analyzing the probabilities of these images. Their findings indicate that the initial model layer is capable of detecting various edge types, while subsequent layers aggregate more detailed features. This hierarchical processing ultimately flattens the information, leading to effective image detection. Their model achieves a notable 96\% accuracy on test images, with training and validation losses at 14.2\% and 3.6\%, respectively, and a validation accuracy of 99.2\%.

Vaibhav Swaminathan et al. \cite{2} expand the scope of autonomous driving by developing a system capable of detecting and recognizing road signs, as well as calculating the distance between the car and these signs. Additionally, they propose a lane-following method to maintain the vehicle within its lane. The system, which combines image recognition capabilities with a CNN and an Arduino-controlled autonomous car, was tested using the Belgium traffic signs dataset. The accuracies achieved by four different models are summarized in Table \ref{tab:model_accuracies}.

\begin{table}[h]
    \centering
    \caption{Accuracies of different models on the Belgium traffic signs dataset.}
    \label{tab:model_accuracies}
    \begin{tabular}{lc}
        \hline
        Model & Accuracy (\%) \\
        \hline
        ACNN4 & 72.1 \\
        PHOG + SVM & 82.01 \\
        CNN-EIM Model & 82.4 \\
        MobileNet CNN & 83.7 \\
        \hline
    \end{tabular}
\end{table}

Mohinsa Binte Asad et al. \cite{3} propose a TSR method based on HSV transformation and template matching, capable of detecting and recognizing traffic signs of various shapes, including circular, triangular, rectangular, and octagonal, covering all existing Bengali road signs. The system operates in two stages: sign localization and recognition, requiring 2.846 seconds for color information processing and 5 to 6 seconds for recognition on average. This method achieved an overall accuracy of 83.3\%, with localization and recognition accuracies of 85.34\% and 82.89\%, respectively.

Pierre Sermanet and Yann LeCum \cite{4} applied ConvNets to traffic sign classification, achieving second-best accuracy in the GTSRB competition with 98.97\%, closely following the best entry at 98.98\% and surpassing human performance. Further experiments increased their record to 99.17\% by enhancing network capacity and using grayscale images.

Chunsheng Liu et al. \cite{5} developed a high-performance TSR framework incorporating a novel region-of-interest extraction method, a split-flow cascade tree detector, and a rapid occlusion robust classification method based on extended sparse representation. This approach rapidly detects and recognizes multiclass traffic signs with high accuracy, achieving a 94.81\% accuracy rate, a 4.10\% false alarm rate, and a 115 ms processing time per frame.

Andreas Mogelmo se et al. \cite{6} extended the research to the detection of U.S. traffic signs, highlighting the variability in sign appearance across different regions. They improved the LISA-TS dataset and applied both Integral Channel Features and Aggregate Channel Features detection methods, demonstrating superior performance on U.S. signs.

Domen Tabernik and Danijel Skocaj \cite{7} focused on automating traffic sign inventory management using the Mask R-CNN for detection and recognition. This approach, tested on 200 traffic sign categories, showed less than 3\% error rates, proving its effectiveness for practical applications.

The collective insights from these studies underscore the diversity of approaches and technologies applied in the pursuit of efficient and accurate traffic sign detection and recognition, reflecting the dynamic nature of research in this domain.

\section{Methodology}\label{sec2}

The principal aim of the proposed system is to classify, recognize, and identify traffic signs by employing Convolutional Neural Networks (CNN). This involves a comprehensive exploration and summarization of each dataset to fully visualize them, followed by the design, training, and testing of the model architecture. Upon completion of these steps, the model is then applied to predict and characterize new images based on the trained dataset, analyzing the maximum likelihoods for the categorization of these new images.

\begin{figure}[!ht]
\centering
\includegraphics[width=\linewidth]{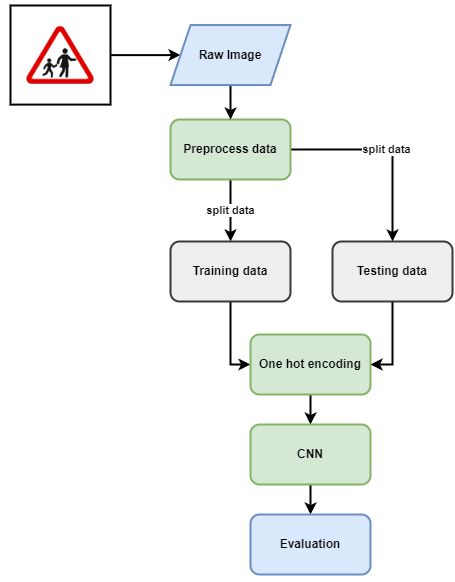}
\caption{Flow diagram representing the system architecture for traffic sign recognition, detailing the steps from raw image acquisition to result generation.}
\label{fig:methodology}
\end{figure}

A CNN is constructed from a series of convolutional and pooling layers, which function collectively to extract salient features from images that best respond to the intended final objective. In the subsequent paragraphs, we delineate the mathematical formalisms that underpin each layer within the network.

The convolutional layer applies a filter to the input image to create feature maps that encapsulate essential aspects of the image. The convolution operation is defined mathematically as follows:

\begin{figure}[H]
\centering
\includegraphics[width=\linewidth]{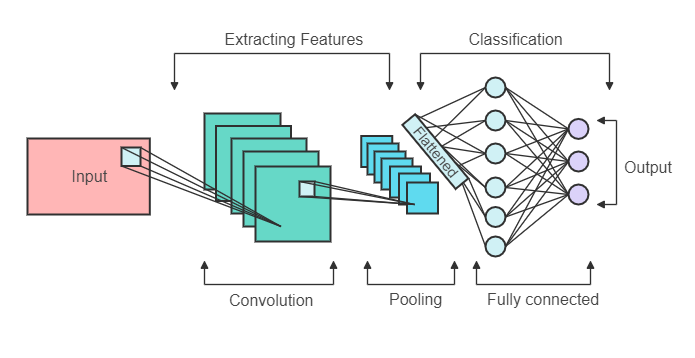}
\caption{Flow diagram of CNN}
\label{fig:cnn}
\end{figure}

\begin{equation}
Conv(I, K)_{x,y} = \sum_{i=1}^{n_H} \sum_{j=1}^{n_W} \sum_{k=1}^{n_C} K_{i,j,k} \cdot I_{x+i-1, y+j-1, k}
\end{equation}
where $I$ represents the input image, $K$ is the filter, and $n_H$, $n_W$, and $n_C$ denote the height, width, and number of channels of the filter, respectively.

The pooling layer reduces the spatial size of the feature maps to decrease the number of parameters and computation in the network. The dimensions of the image after pooling are calculated by:
\begin{equation}
\begin{split}
\text{dim}(\text{pooling}(\text{image})) = \Bigg( &\left\lfloor \frac{n_H + 2p - f}{s} \right\rfloor + 1, \\
&\left\lfloor \frac{n_W + 2p - f}{s} \right\rfloor + 1, n_C \Bigg)
\end{split}
\end{equation}
where \( p \) represents the padding, \( f \) is the filter size, \( s \) is the stride, and \( n_C \) is the number of channels.

The dropout layer helps prevent overfitting by randomly setting a fraction of input units to 0 at each update during training time. The expected output of a neuron after applying dropout is given by:
\begin{equation}
E_R = \frac{1}{2} \left( \sum_{i=1}^{n} p_i w_i I_i \right)^2 + \sum_{i=1}^{n} p_i(1 - p_i) w_i^2 I_i^2
\end{equation}
where $p_i$ denotes the dropout probability for the $i$-th neuron, $w_i$ is the weight, and $I_i$ is the input to the neuron.

The fully connected layer integrates the features learned by the previous layers to perform the final classification. The value for each neuron in the fully connected layer is computed as follows:
\begin{equation}
z_j^{[i]} = \sum_{l=1}^{n_{i-1}} w_{j,l}^{[i]} a_l^{[i-1]} + b_j^{[i]}
\end{equation}
Where \( w_{j,l}^{[i]} \) denotes the weight connecting the \( l \)-th neuron in the \( (i-1) \)-th layer to the \( j \)-th neuron in the \( i \)-th layer, \( a_l^{[i-1]} \) is the activation of the \( l \)-th neuron in the \( (i-1) \)-th layer, and \( b_j^{[i]} \) is the bias term for the \( j \)-th neuron in the \( i \)-th layer.

The methodology for developing a Traffic Sign Recognition System using Convolutional Neural Networks (CNN) encompasses several critical stages:

\subsection{Data Pre-processing}
Utilizing a Jupyter Notebook, the dataset was imported and underwent an initial preprocessing stage. Images, which varied in dimensions, were resized to a uniform \(30 \times 30\) pixel resolution. This uniformity is crucial as it ensures consistency in the input dimensionality across the dataset, facilitating more effective learning by the CNN. The list of images and their corresponding labels were converted into numpy arrays. Class labels were converted into one-hot encoding vectors using the to\_categorical utility from Keras, which simplifies the network's output interpretation for the classification task.

\subsection{Convolutional Neural Network}
The architecture of the CNN was designed to effectively extract and learn the most critical features from the traffic sign images. It consisted of four convolutional layers, with the first two layers having 32 filters of size $5 \times 5$ and the next two layers having 64 filters of size $3 \times 3$. Each convolutional layer used the ReLU activation function to introduce non-linearity, allowing the model to learn complex patterns. To reduce overfitting, dropout layers were applied after pooling layers, with dropout rates of 0.25 after the first two convolutional layers and 0.5 before the final dense layer. The architecture includes two MaxPooling layers designed to downsample the feature maps, effectively reducing their dimensions and the overall computational load. The culmination of the network architecture is a dense layer consisting of 256 units, followed by a softmax classifier with 43 units, each representing a distinct traffic sign category.

\begin{figure}[!ht]
\centering
\includegraphics[width=\linewidth]{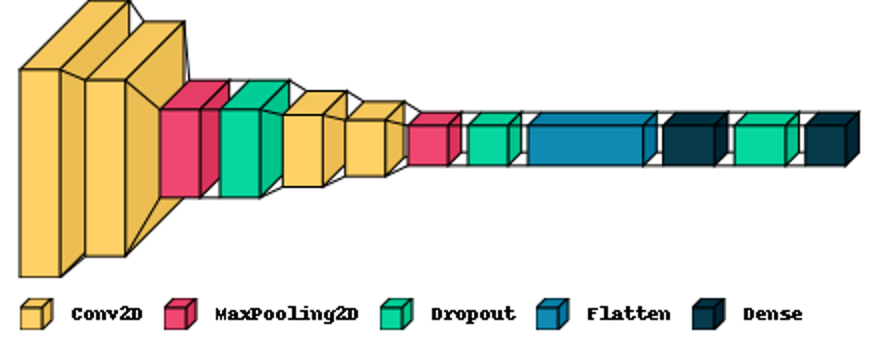}
\caption{Representation of the model}
\label{fig:LossvsEpochs}
\end{figure}

\subsection{Modelling the Data}
The model was compiled using the Adam optimizer, which is known for its efficiency in handling sparse gradients and adaptively tuning itself. The loss function used was categorical crossentropy, ideal for multi-class classification problems. An initial learning rate of 0.001 was set, with a dynamic reduction strategy that decreased the learning rate if the validation accuracy did not improve, effectively aiding in the model's convergence.

\subsection{Training and model evaluation}
The model was trained using mini-batch gradient descent with a batch size of 32 for 100 epochs. To prevent overfitting and to ensure the model's generalizability, an EarlyStopping callback was used, which monitored the validation accuracy and stopped training if no improvement was observed for 10 consecutive epochs. The training process was visualized through the plotting of accuracy and loss graphs for both training and validation sets. These graphs depicted a swift increase in learning performance in the initial epochs, followed by stabilization, illustrating both the model's rapid learning capability and its generalization performance on unseen data.

\begin{figure}[!ht]
\centering
\includegraphics[width=\linewidth]{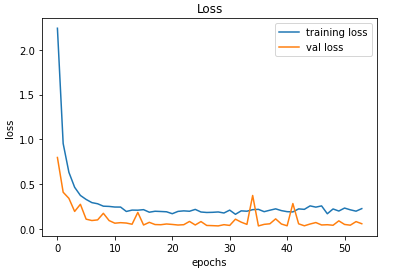}
\caption{Training and validation loss curves over 50 epochs, illustrating the model's learning progression.}
\label{fig:LossvsEpochs}
\end{figure}

\subsection{Output}
The model's predictive performance was systematically evaluated. A confusion matrix was constructed to visualize the accuracy across all classes, revealing both high-performing and underperforming categories. This diagnostic tool was instrumental in identifying patterns of misclassification, offering insights for further refinement.

\begin{figure}[!ht]
\centering
\includegraphics[width=\linewidth]{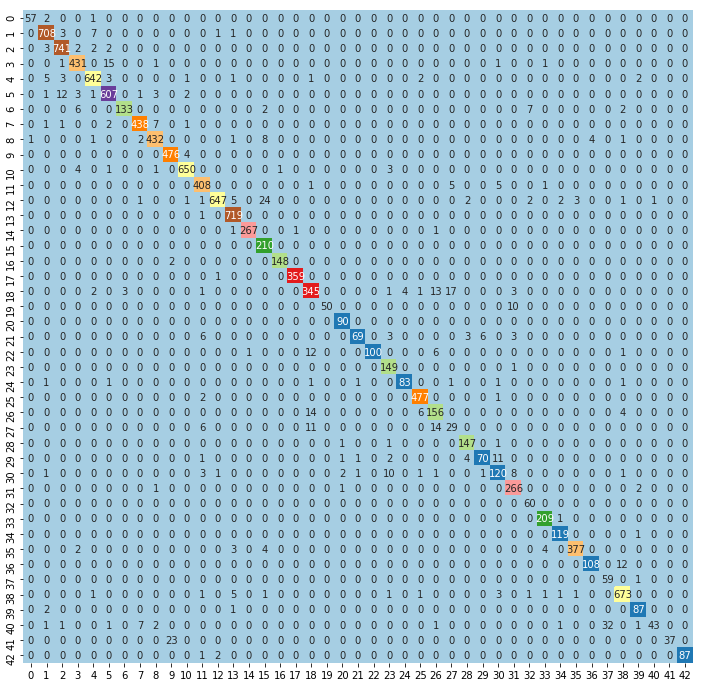}
\caption{Confusion matrix showcasing the model's performance in accurately classifying traffic sign images.}
\label{fig:Matrix}
\end{figure}

\begin{figure}[H]
\centering
\includegraphics[width=\linewidth]{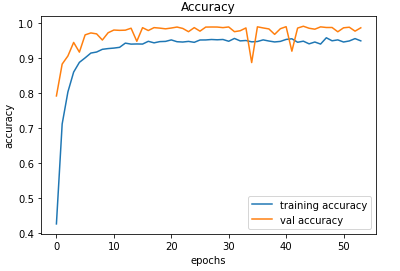}
\caption{Training and validation accuracy curves over 50 epochs, showing the model's performance improvement and stabilization.}
\label{fig:AccuracyVsEpochs}
\end{figure}

\section{Results and Analysis}

The proposed Convolutional Neural Network (CNN) was rigorously evaluated using the German Traffic Sign Recognition Benchmark (GTSRB) dataset. This dataset comprises 43 distinct classes of traffic signs, catering to various road scenarios and regulations, such as speed limits, prohibitory, mandatory, and warning signs. The dataset, consisting of 39,209 images, was partitioned into a training set of 31,367 images and a testing set of 7,842 images to assess the model's performance.

The CNN's architecture was tailored to handle the intricacies of traffic sign images, yielding a high training accuracy of 94.76\%. Various batch sizes were tested, with the larger batch size of 64 resulting in superior performance. Notably, the model's accuracy plateaued at around 40 epochs, suggesting the limits of learning under the given architecture and training data were reached.

\begin{figure}[!ht]
\centering
\includegraphics[width=\linewidth]{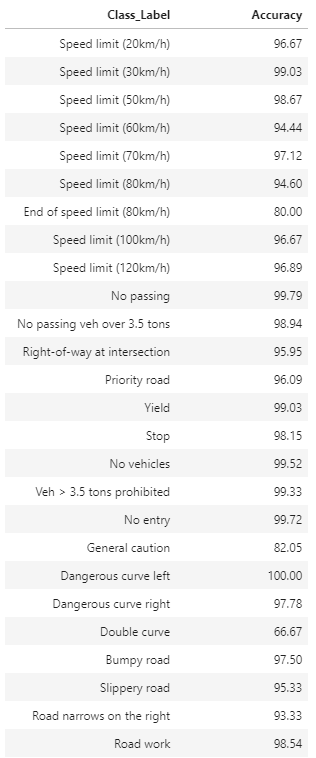}
\caption{Class-wise accuracy chart showing the model's performance across different traffic sign categories.}
\label{fig:ClassWiseAccuracy}
\end{figure}

\begin{figure}[!ht]
\centering
\includegraphics[width=\linewidth]{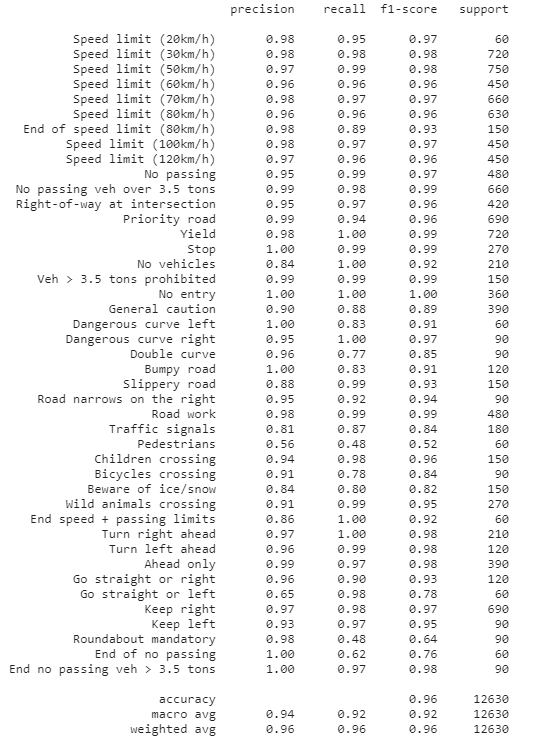}
\caption{Detailed classification report providing precision, recall, f1-score, and support for each traffic sign class recognized by the model.}
\label{fig:ClassificationReport}
\end{figure}

Upon evaluation with the testing set, the CNN demonstrated a robust generalization capability with a commendable accuracy of 95.85\%. The progression of accuracy and loss over epochs is depicted in Figure \ref{fig:LossvsEpochs}, highlighting consistent improvement and stabilization, a testament to the CNN's predictive performance and learning efficiency.

Class-wise accuracy rates are illustrated in Figure \ref{fig:ClassWiseAccuracy}, showcasing the model's variable performance across different traffic sign categories. Notably, signs such as 'Speed limit (20km/h)' and 'Double curve' indicated a need for model refinement, possibly through data augmentation or architectural adjustments to improve recognition in underrepresented classes.

A classification report was generated to give a more granular view of the model's performance, offering precision, recall, f1-scores, and support for each class (Figure \ref{fig:ClassificationReport}). This report indicates high precision and recall for most classes, with some exceptions that highlight potential areas for further research and improvement.

\section{Conclusion}
This study detailed the development and validation of a CNN-based Traffic Sign Recognition System. Through methodical experimentation and analysis, the system demonstrated high accuracy in classifying various traffic signs, outperforming benchmarks set by prior models. These results affirm the potential of CNNs in real-world applications for traffic sign recognition, particularly in enhancing road safety and supporting the advancement of autonomous vehicle technologies. Future work may explore the integration of more complex models or additional data sources to address the minor classification challenges observed in specific sign categories.

\end{document}